# LLM-Agent-Controller: A Universal Multi-Agent Large Language Model System as a Control Engineer

Rasoul Zahedifar[1,2], Sayyed Ali Mirghasemi[1], Mahdieh Soleymani Baghshah[2], Alireza Taheri[1,3,*]

[1] Social and Cognitive Robotics Laboratory, Mechanical Engineering Department, Sharif University of Technology, Tehran, Iran

[2] Machine Learning Laboratory, Computer Engineering Department, Sharif University of Technology, Tehran, Iran

[3] Artificial Intelligence in Design and Complex Systems (AIDACS) Group, Sharif University of Technology, Tehran, Iran

[*] Corresponding author: artaheri@sharif.edu , Tel: +982166165531

**Abstract**

This study presents the LLM-Agent-Controller, a multi-agent large language model (LLM) system developed to address a wide range of problems in control engineering (Control Theory). The system integrates a central controller agent with multiple specialized auxiliary agents, responsible for tasks such as controller design, model representation, control analysis, time-domain response, and simulation. A supervisor oversees high-level decision-making and workflow coordination, enhancing the system's reliability and efficiency. The LLM-Agent-Controller incorporates advanced capabilities, including Retrieval-Augmented Generation (RAG), Chain-of-Thought reasoning, self-criticism and correction, efficient memory handling, and user-friendly natural language communication. It is designed to function without requiring users to have prior knowledge of Control Theory, enabling them to input problems in plain language and receive complete, real-time solutions. To evaluate the system, we propose new performance metrics assessing both individual agents and the system as a whole. We test five categories of Control Theory problems and benchmark performance across three advanced LLMs. Additionally, we conduct a comprehensive qualitative conversational analysis covering all key services. Results show that the LLM-Agent-Controller successfully solved 83% of general tasks, with individual agents achieving an average success rate of 87%. Performance improved with more advanced LLMs. This research demonstrates the potential of multi-agent LLM architectures to solve complex, domain-specific problems. By integrating specialized agents, supervisory control, and advanced reasoning, the LLM-Agent-Controller offers a scalable, robust, and accessible solution framework that can be extended to various technical domains.

1. **Introduction**

In recent years, researchers have leveraged large language models (LLMs) as foundational tools for developing AI agents, resulting in significant advancements in the field [1-4]. These agents are now being deployed across a wide range of tasks, such as code generation [5], data analysis [6], content creation [7, 8], and complex decision-making systems [9-13]. As these LLM agents become more capable, they are transforming industries and reshaping the way we interact with technology. However, challenges remain in the field of control engineering process where ensuring stability, robustness, efficiency, and performance of the control workflow is critical.

Control engineering, a field focused on the control design and analysis of dynamic systems, often requires intricate problem-solving across a variety of specialized areas, from controller design to system simulation [14, 15]. Traditionally, this field has relied heavily on expert knowledge and sophisticated software tools to solve complex control problems. However, the increasing complexity of modern systems, coupled with the growing need for accessible solutions, has highlighted the limitations of existing approaches, particularly in terms of user accessibility and integration of interdisciplinary knowledge [16, 17]. To address this gap, we propose a novel solution that leverages the capabilities of an LLM multi-agent system, offering a more intuitive, integrated, and efficient approach to tackling control engineering challenges.

This research presents the LLM-Agent-Controller, a robust multi-agent system powered by LLMs. As its director, a supervisor LLM acts as the central decision-maker, orchestrating a network of specialized LLM agents to optimize task execution and workflow efficiency. These agents function within an efficient architecture, ensuring seamless interaction, effective task distribution, and streamlined operations. At its core, the controller LLM agent is capable of handling a wide range of challenges, including controller design, model representation, control analysis, and simulations. Additionally, the LLM-Agent-Controller leverages a diverse set of auxiliary LLM agents, each designed to handle specialized tasks such as retrieving information from external data sources, facilitating personalized interactions, and enhancing overall system adaptability and versatility. By distributing responsibilities across dedicated agents focused on planning, retrieval, research, reasoning, critique, debugging, memory management, and communication, the system ensures seamless operation and optimized performance.

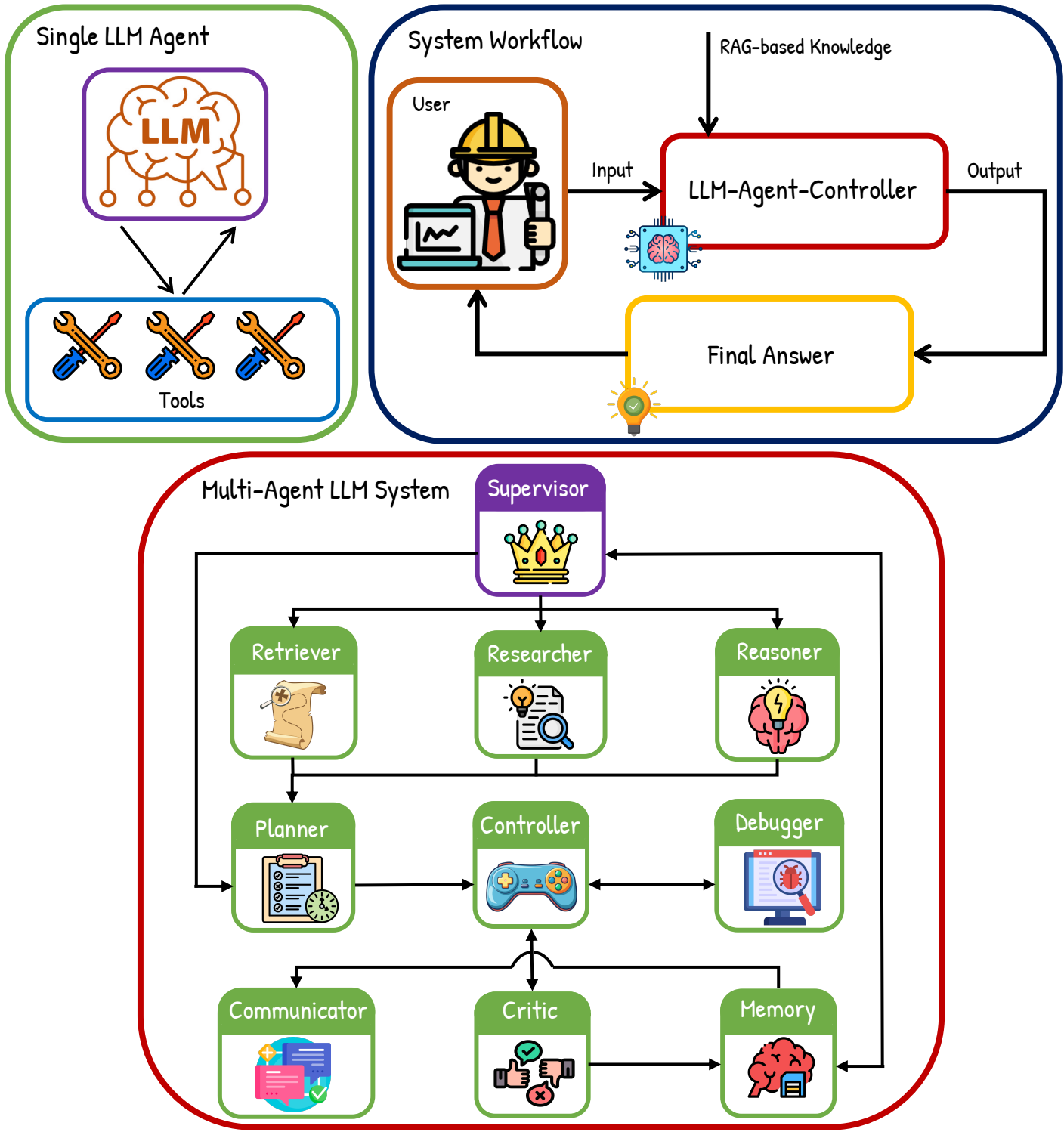

Figure 1. Workflow of the proposed LLM-Agent-Controller system with collaborative multi-agent coordination.

To the best of our knowledge, this is the first framework designed to automate and enhance controller design and stability analysis for control systems and robotics. Traditionally, engineers manually designed controllers tailored to specific tasks and requirements. With the LLM-Agent-Controller, the entire controller design process can now be managed autonomously by an interconnected network of LLM agents. Users simply submit queries and receive real-time responses, enabling an interactive and iterative problem-solving experience. The workflow of the proposed LLM-Agent-Controller is depicted in Figure 1.

The key contributions of the proposed LLM-Agent-Controller are as follows:

**1. A Universal LLM Agent for Control Theory:** The proposed LLM-Agent-Controller is the first powerful, universal multi-agent LLM system tailored specifically for Control Theory. Unlike prior works with limited functionality, our system offers an extensive suite of tools to address a wide range of tasks. These include controller design tailored to user needs and systems, modeling and representation of control systems, time- and frequency-domain simulations, and stability analysis.

**2. A Multi-agent Framework to Enhance Adaptability and Versatility:** Our proposed framework introduces a novel architecture, where a range of specialized LLM agents collaborate to handle various tasks under the management of a central supervisor LLM. Unlike traditional single-agent controller models, this multi-agent approach fosters an optimal, robust workflow, offering enhanced adaptability and versatility in response to user requests.

**3. Retrieval-Augmented Generation Knowledge Support:** The proposed framework utilizes Retrieval-Augmented Generation (RAG) to integrate external knowledge bases (provided by the user), ensuring accurate, context-aware, and comprehensive responses. By retrieving relevant information from trusted sources, the system delivers reliable and well-informed answers tailored to user queries.

**4. Advanced Reasoning and Workflow Optimization for Enhanced Decision-Making:** The proposed LLM-Agent-Controller integrates advanced reasoning capabilities and workflow optimization to deliver intelligent, context-aware solutions. It intelligently retrieves real-time information from external sources on the Internet, adapts workflows to enhance efficiency, and applies sophisticated reasoning techniques, such as Chain-of-Thought (CoT) and Tree-of-Thought (ToT), to refine decision-making processes. This dynamic approach ensures that complex tasks are handled with precision and robustness, ultimately enabling the system to generate reliable and highly accurate results tailored to the user's needs.

**5. Self-correcting Mechanism for Enhanced Robustness:** The proposed system automatically verifies the accuracy of its responses before delivering the final answer to the user. Additionally, it is designed with built-in robustness, allowing it to detect and debug errors or bugs that may arise during execution, ensuring reliable performance.

**6. End-to-End Interaction with User-Friendly Experience:** The LLM-Agent-Controller is designed for effortless usability, allowing users to ask questions in natural language without needing prior expertise in Control Theory. The framework also features memory capabilities, enabling it to recall past conversations, user priorities, and personalized preferences for a more tailored experience. Furthermore, communication between the user and the system can be adapted to different modes (voice, text), formats (PDF, plain text), and languages, ensuring flexibility and accessibility.

The structure of this paper is organized as follows: Section 2 provides a review of related work on LLM agents in robotics and control systems. Section 3 details the methodology of the proposed LLM-Agent-Controller system and introduces performance metrics to assess the algorithm's effectiveness across different scenarios. Section 4 presents the results and analysis of the numerical simulations. Section 5 discusses the limitations of the current approach and outlines potential directions for future research. Finally, the paper concludes in Section 6. The workflow of the proposed LLM-Controller scheme is illustrated in Figure 1. The colors used in the figures throughout the paper are intentionally chosen to be meaningful and visually harmonious.

2. Background

In this section, some of the relevant works are reviewed to provide a comprehensive understanding of the existing methods and advancements in control systems, particularly in the context of LLM agent-based approaches.

Keivan et al [18] evaluated the abilities of LLMs like GPT-4, Claude 3 Opus, and Gemini 1.0 Ultra in solving undergraduate-level control engineering problems using a new benchmark dataset called ControlBench. The dataset captures the complexity of classical control design, and expert evaluations reveal that Claude 3 Opus outperforms the others, though all models struggle with visual tasks like Bode and Nyquist plots. A simplified version, ControlBench-C, is introduced for faster, automatic evaluations, but lacks the depth of the full dataset. The study provides valuable insights into the strengths and limitations of current LLMs in control engineering. However, by default, LLMs are not tailored for control engineering tasks and may face challenges when addressing such problems.

Gou et al [19] introduced ControlAgent, an automated framework for control system design that integrates LLMs with domain-specific expertise. ControlAgent uses multiple collaborative agents, including LLMs and a Python computation agent, to iteratively refine controller parameters based on user-specified requirements for stability, performance, and robustness. To evaluate its effectiveness, the authors develop ControlEval, a benchmark dataset of 500 diverse control tasks, and demonstrate that ControlAgent outperforms traditional toolbox-based methods and other LLM-based approaches. By mimicking human iterative design processes, ControlAgent offers a fully automated, end-to-end solution for control engineering tasks. This work represents a significant advancement toward automating complex engineering workflows. However, ControlAgent possesses too few tools and capabilities when navigating the broad spectrum of challenges in control engineering.

Zahedifar et al [20] presented the LLM-Controller, a novel approach for dynamic adaptation in robot controllers using LLMs. The proposed controller automatically adjusts to changes in system dynamics or reference signals, demonstrating robust performance across various real-world scenarios without manual tuning. By integrating LLM reasoning capabilities with a nonlinear controller, the method enhances adaptability, versatility, and stability in uncertain environments. Evaluation of 2-link and 3-link robotic manipulators shows high success rates and efficiency, further enhanced through the use of CoT reasoning. The study highlights the LLM-Controller's flexibility, generalizability, and potential as a powerful tool for dynamic and complex robotic systems.

Ni et al [21] developed MechAgents, a novel framework that uses collaborative LLMs to solve mechanics problems, particularly in elasticity, by integrating physics-inspired modeling with AI-driven automation. By leveraging the diverse capabilities of multiple AI agents, the framework automates tasks like retrieving knowledge, writing and executing code, and analyzing results for problems involving finite element method (FEM). For simple tasks, a two-agent team successfully handles code generation, execution, and self-correction, while more complex problems are addressed by a larger group of agents with specialized roles such as planning, formulation, and criticism. The agents dynamically collaborate and mutually refine their outputs, enhancing the accuracy and reliability of solutions. This approach combines the adaptability of LLMs with the robustness of physics-based modeling, demonstrating the potential for automating complex engineering problem-solving processes through intelligent agent teamwork.

Jadhav et al [22] investigated a novel framework that integrates pre-trained LLMs with a FEM module to revolutionize mechanical design by automating the iterative optimization process. Unlike conventional methods that rely on expert-driven refinements or machine learning models demanding extensive training, this approach allows LLMs to reason, plan, generate, and optimize designs using FEM feedback without domain-specific training. The framework demonstrates its effectiveness by achieving a 90% success rate in generating and refining truss structure designs based on natural language specifications, with success rates varying by constraint complexity. Employing prompt-based optimization, LLMs iteratively improve designs by reasoning through solution-score pairs, showcasing their ability to autonomously refine solutions. This integration of LLMs and FEM ensures efficient, accurate, and adaptive design processes, marking a transformative advancement in automated engineering solutions.

Lu et al [23] explored the potential of using LLMs in mechanical design by proposing a method to construct a comprehensive Mechanical Design Agent (MDA) through guided LLM learning. While directly using LLMs for mechanical design tasks is currently ineffective, the approach demonstrates that with proper guidance, LLMs can achieve modeling capabilities similar to professional designers in certain areas. Experimental results show promising outcomes, although MDAs still require substantial learning and guidance to become experts across all aspects of mechanical design. The research highlights that while LLMs can enhance productivity, significant work remains before they can fully replace human designers. This study provides valuable insights and methods for advancing the use of LLMs in mechanical design, underscoring both the opportunities and challenges ahead.

Building upon the findings from the literature review, the following section presents the methodology employed to fill the identified gaps and push the research forward. It describes the design and development of the LLM-Agent-Controller, along with the approaches used to evaluate its adaptability and performance in various problems and tasks.

## 3. Methodology

In this section, we introduce the proposed LLM-Agent-Controller system as an advanced framework for solving a wide range of control engineering (Control Theory) problems. This system extends our previous work, LLM-Controller [20]. The system allows users to ask questions in natural language and receive real-time responses. This interaction continues seamlessly until the user initiates a new topic. The LLM-Agent-Controller consists of multiple agents working together to enhance robustness, versatility, adaptability, flexibility, and accessibility. We begin by describing the system, starting with its network graph and workflow performance. Next, we outline the tools and capabilities of each individual agent, and as a result, the integrated system as a whole. Finally, we introduce novel metrics designed to evaluate the performance and capabilities of the proposed system.

### 3.1. Network Graph

The LLM-Agent-Controller system is composed of nine agents, along with a Supervisor LLM positioned at the highest level to manage the workflow. At the core of the system, the Controller Agent is responsible for generating answers related to the user's question, while other agents support the process in various ways. These agents are coordinated and managed by the Supervisor, which is an LLM, not an agent, and oversees the operation, as illustrated in Figure 2.

When a user submits a query, it is first received by the Supervisor, who acts as the decision-maker. Depending on the nature of the query, the Supervisor determines whether assistance is needed from additional agents; The Retriever Agent utilizes the RAG technique [24] to efficiently fetch specific data from external resources provided by the user, the Researcher Agent can search the internet for relevant information, and the Reasoner Agent can apply logical reasoning techniques including CoT [25, 26] or ToT [27, 28], to enhance the quality of the response. If the answer may already exist in past interactions, the Supervisor can call the Memory Agent to retrieve the relevant information efficiently. If no external assistance is required, the Supervisor passes the query to the Planner Agent, which organizes the sequence of functions (or tools) that the Controller Agent will use to generate the response. Once the planning is complete, the Controller Agent, positioned at the heart of the system, executes the necessary functions to construct the answer. If any errors or bugs arise during this process, the Controller Agent consults the Debugger Agent for troubleshooting guidance before proceeding. Once an answer is generated, the Critic Agent, a trusted judge [29], reviews it to ensure alignment with the original user query. If the answer does not meet the necessary criteria, the Critic

Agent provides feedback for improvement, prompting the Controller Agent to refine the response. If the answer is validated, it is stored efficiently by the Memory Agent for potential future use. Before delivering the response to the user, the Communicator Agent ensures that it is presented in the requested mode, format, or language. If no specific formatting is required, the system displays the answer directly to the user.

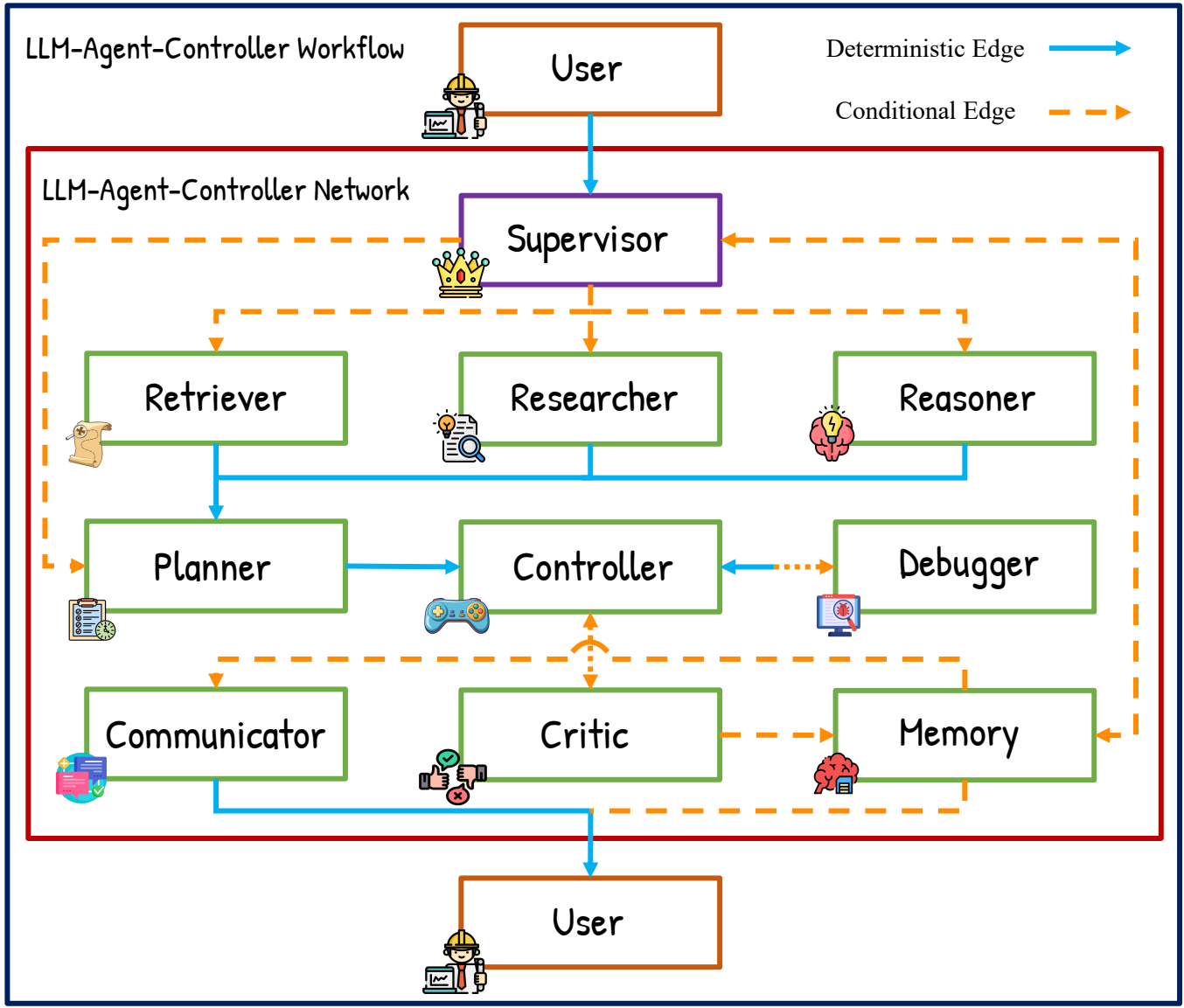

Figure 2. The architecture of the LLM-Agent-Controller system framework.

As depicted in Figure 2, the system is structured as a network where agents and the Supervisor function as interconnected nodes. When only one agent is responsible for handling the next step, the connection follows a deterministic path. In cases where multiple agents could be involved, the system follows a conditional approach, dynamically determining the most suitable agent based on the context of the user query, agent responsibilities, and available tools.

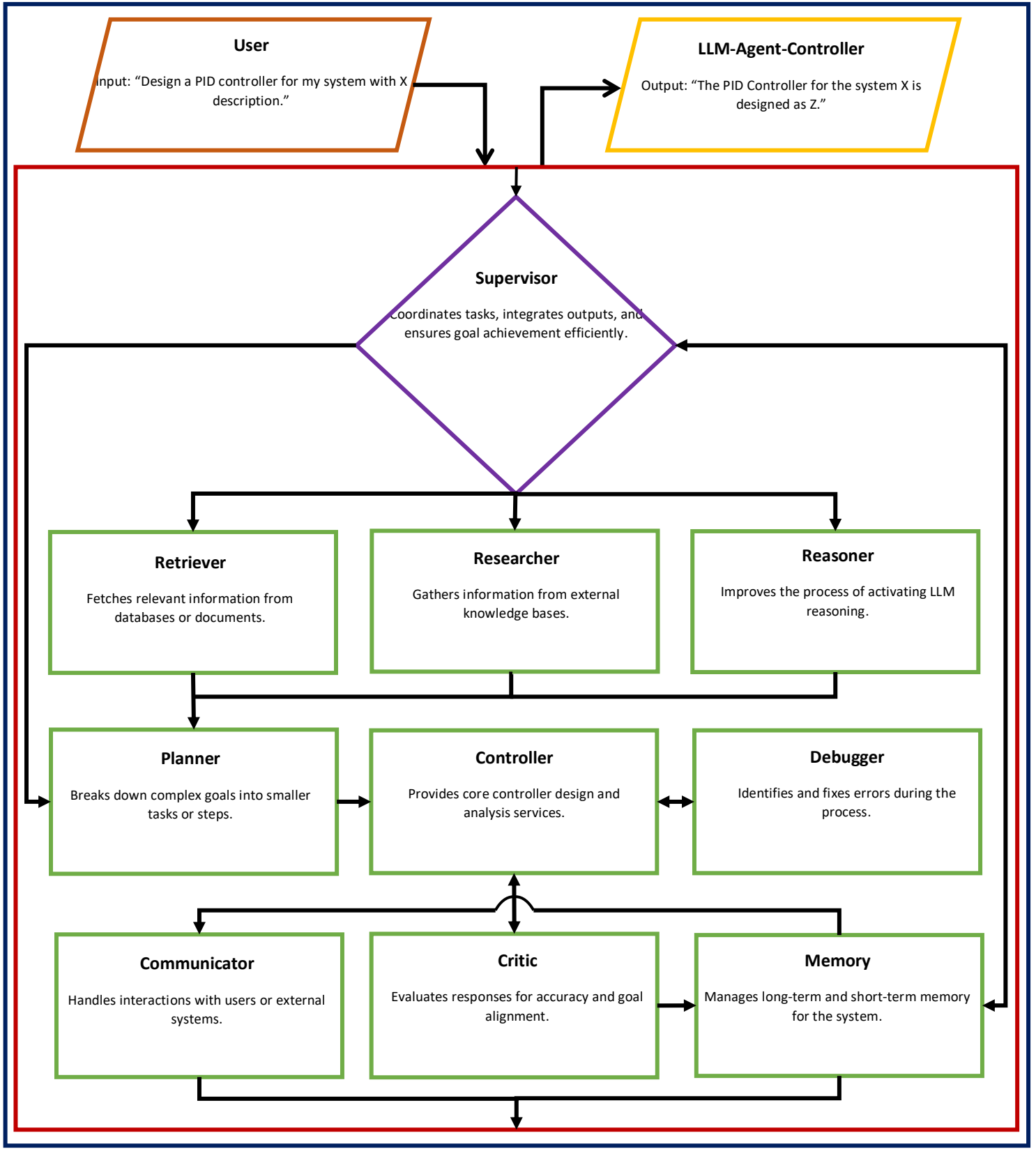

Figure 3. A schematic representation of the LLM-Agent-Controller system.

The LLM-Agent-Controller system is designed to maintain an ongoing, intelligent conversation with the user, efficiently managing complex control tasks by distributing responsibilities among specialized agents. The role of each node in the network is further illustrated in Figure 3. This structured yet flexible approach ensures a smooth and effective problem-solving experience, allowing the system to adapt dynamically to different user needs.

### 3.2. Execution Tools and Capabilities

Unlike previous works on LLM agents, where the agent (or the primary agent in a multi-agent system) has access to a limited set of tools, our proposed LLM-Agent-Controller features a Controller Agent at its core, equipped with more than 140 tools.

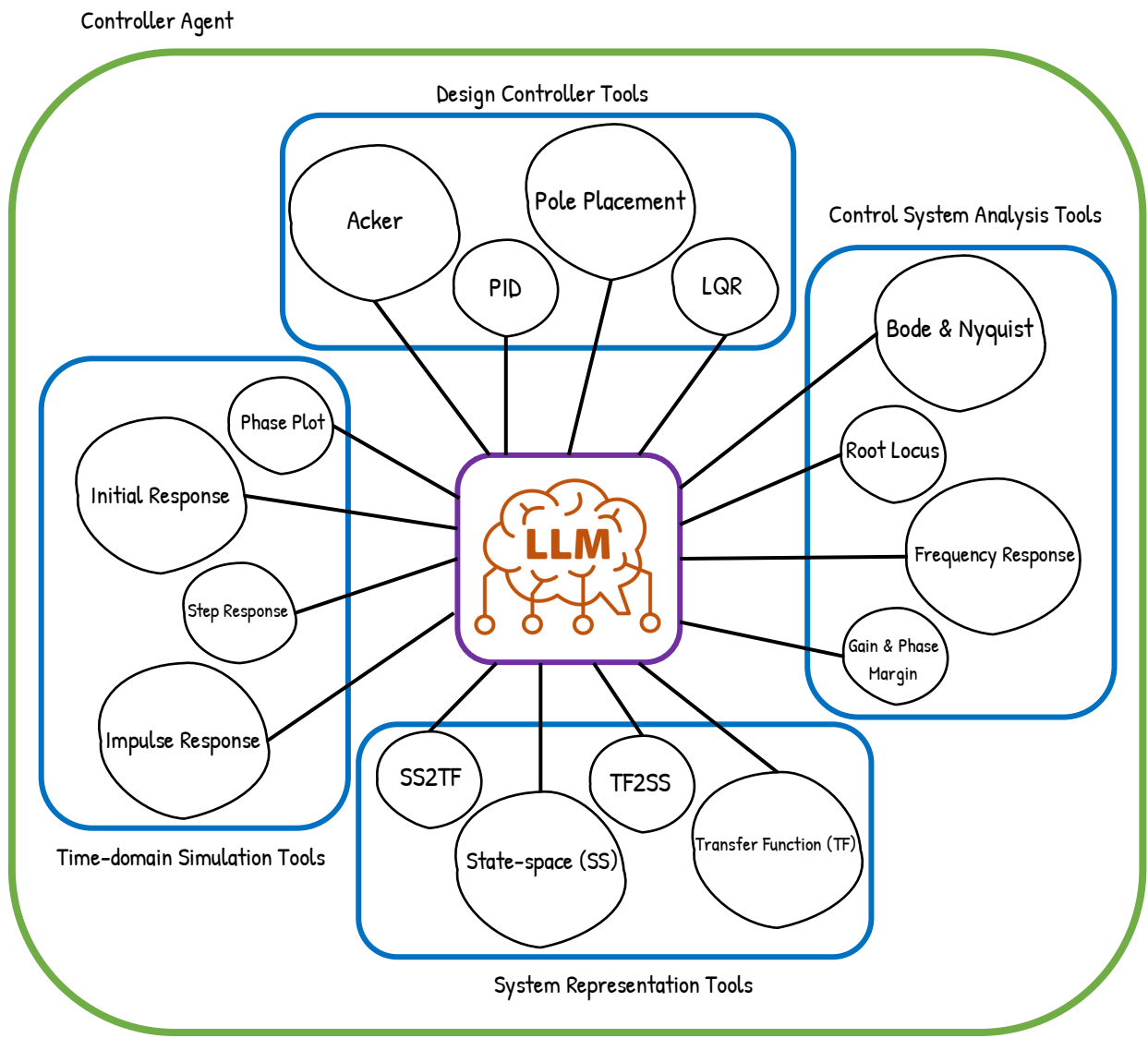

Figure 4. A schematic representation of the controller agent with its tools.

This significantly expands the agent's capabilities, addressing a key limitation of conventional LLM agents, namely, their reliance on pre-implemented tools designed by human developers.

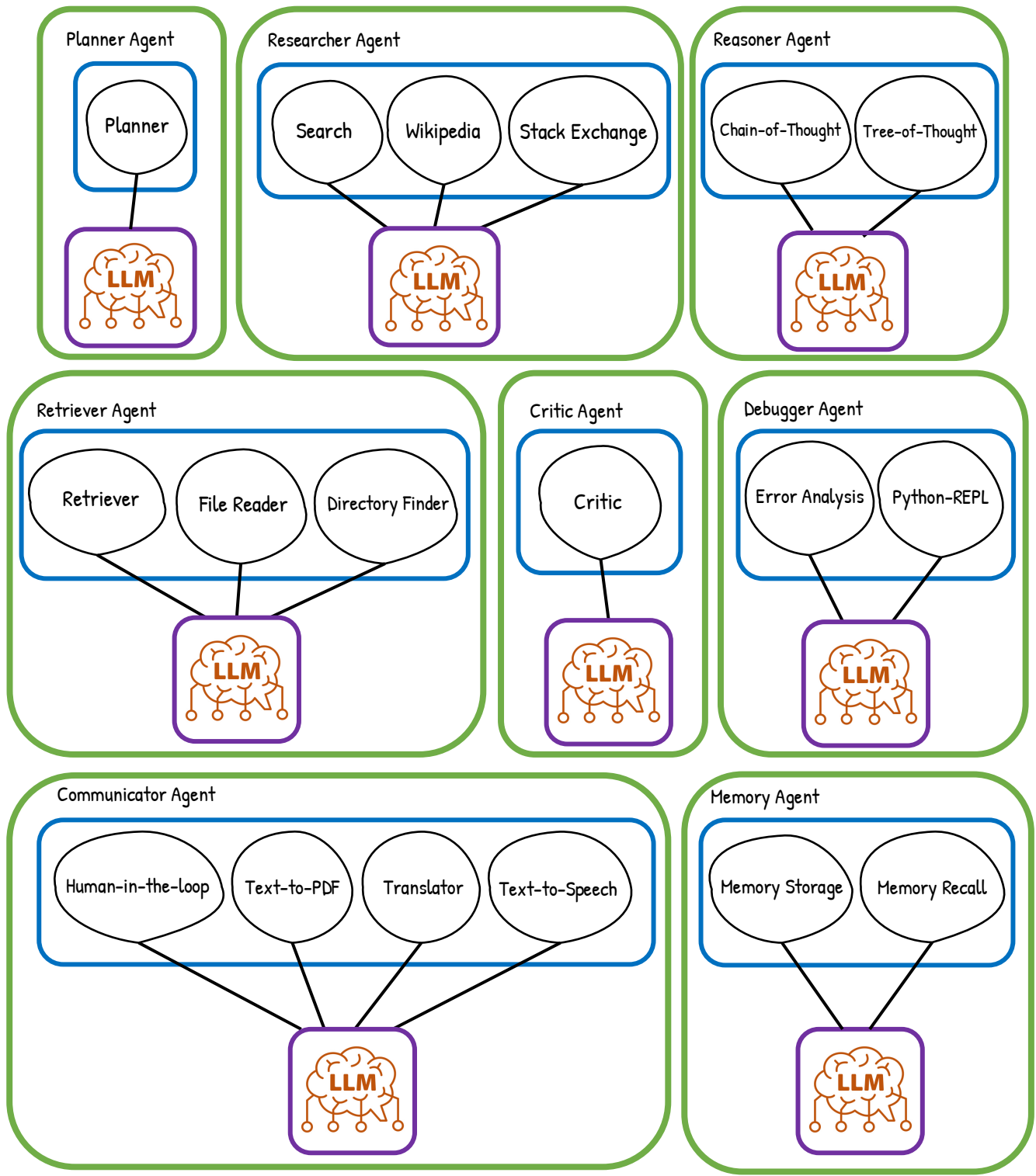

Figure 5. A schematic representation of the auxiliary agents with their tools.

Such constraints hinder their applicability in domains requiring a vast array of highly specialized tools, such as life sciences, engineering, and medicine [30]. Moreover, human-designed tools are typically confined to existing expert-developed solutions, while AI-generated tools, though promising, often come with efficiency limitations [31-33]. To address this limitation, we propose using existing library packages as a comprehensive tools package, leveraging the vast number of specialized libraries available. However, to prevent token limitations in LLMs, we must restrict the number of functions included or provide compact descriptions of them [34], since a large number of tools might cause token limitations in LLMs.

One such library integrated into our system is the Python Control Library [35], which provides functions for linear input/output systems in state-space and frequency domains, block diagram algebra, time response analysis, frequency response analysis, estimator design, and nonlinear systems. We categorize these functions, serving as tools for the Controller Agent, into four categories: system representation, controller analysis, controller design, and simulation, with key examples illustrated in Figure 4.

In addition to the Controller Agent, the proposed system incorporates auxiliary agents to improve robustness, versatility, adaptability, flexibility, and accessibility. Figure 5 presents the tools available to each agent. Appendix A provides all the prompts and pseudocode for the proposed LLM-Agent-Controller workflow. However, prompt engineering could further enhance the model's performance [36], and this will be explored in future work.

### 3.3. Metrics

In addition to the output of the LLM-Agent-Controller and its simulation plots, which provide a qualitative representation of this work, we establish several metrics to quantitatively evaluate various performance aspects. These scores range from 0 to 1, with higher values indicating better performance of the metric. In the following, we present ten evaluation metrics: the first eight are designed to assess the performance of specific agents, while the final two measure the overall performance of the entire system.

#### 3.3.1. Efficiency Score

The "Efficiency Score" evaluates Controller Agent effectiveness in providing the correct response as the primary part of our final answer. It is defined as follows:

$$\mathcal{M}_E = \frac{1}{\tau}\sum_{t=1}^{\tau} \frac{1}{N}\sum_{n=1}^{N} \mathbb{I}(Correct\ Answer) \tag{1}$$

where $\tau$ is the total number of runs of the LLM-Agent-Controller system, $N$ is the total number of times the Controller Agent is called during a single run of the system, and $\mathbb{I}(\cdot)$ is an indicator function that returns 1 if the specified condition is met and 0 otherwise. This metric provides insight into each agent's performance and effectiveness. By identifying and potentially eliminating less critical agents, the system can be optimized for speed and efficiency.

### 3.3.2. Routing Score

The "Routing Score" assesses whether an agent correctly routes to the appropriate next agent. It is defined as:

$$\mathcal{M}_R = \frac{1}{\tau}\sum_{t=1}^{\tau}\frac{1}{Q}\sum_{q=1}^{Q} \mathbb{I}(Routed\ Next\ Agent_q = Correct\ Next\ Agent_q) \quad (2)$$

where $\mathbb{I}(\cdot)$ equals 1 if the conditional edge to the next agent is chosen correctly and 0 otherwise; and $Q$ denotes the total number of routing decisions to the next agent made during a single run. This metric does not apply to agents with deterministic transitions to their next agents.

### 3.3.3. Arrangement Score

The "Arrangement Score" measures the system's performance in utilizing the optimal arrangement of agents to obtain a response. The score is calculated as follows:

$$\mathcal{M}_A = \frac{1}{\tau}\sum_{t=1}^{\tau}\frac{1}{Q}\sum_{q=1}^{Q} \mathbb{I}(Executed\ Agent_q = Arranged\ Agent_q) \quad (3)$$

where $\mathbb{I}(\cdot)$ equals 1 if the executed agents in the system are the most optimized, appropriately arranged to produce the response, and 0 otherwise; and $Q$ denotes the total number of agents utilized during a single run. This score emphasizes the importance of calling the correct agents in the optimal sequence, neither more nor less, to achieve the desired response efficiently.

### 3.3.4. Planning Score

The "Planning Score" evaluates the performance of the Planner Agent in developing the optimal plan for the Controller Agent. It is defined as follows:

$$\mathcal{M}_P = \frac{1}{\tau}\sum_{t=1}^{\tau}\frac{1}{N}\sum_{n=1}^{N} \mathbb{I}(Planned\ Tools = Correct\ Tools) \quad (4)$$

where $\mathbb{I}(\cdot)$ equals 1 if the tools planned by the Planner Agent match the tools required by the Controller Agent to execute and achieve the response, and 0 otherwise; and $N$ denotes the total number of times the Planner Agent is invoked during a single run.

### 3.3.5. Judgement Score

The "Judgement Score" assesses the performance of the Critic Agent in verifying the output of the Controller Agent. This score is determined as follows:

$$\mathcal{M}_J = \frac{1}{\tau}\sum_{t=1}^{\tau}\frac{1}{N}\sum_{n=1}^{N}[\mathbb{I}(Correct\ Answer = Accepted\ Answer) \\ \vee\ \mathbb{I}(Incorrect\ Answer = Unaccepted\ Answer)] \quad (5)$$

where $\vee$ represents the logical OR function, and $\mathbb{I}(\cdot)$ equals 1 if the Critic Agent correctly verifies a correct answer or denies an incorrect answer, and 0 otherwise; and $N$ is the total number of times the Critic Agent is activated throughout a single execution of the system.

### 3.3.6. Self-Correcting Score

The "Self-Correcting Score" measures the performance of the Debugger Agent in detecting and fixing errors when a bug or issue arises during the execution of the Controller Agent. It is defined as follows:

$$\mathcal{M}_S = \frac{1}{\tau}\sum_{t=1}^{\tau}\frac{1}{N}\sum_{n=1}^{N}\left[\frac{1}{2}\bigl(\mathbb{I}(Error\ Detection) + \mathbb{I}(Error\ fixing)\bigr)\right] \quad (6)$$

where $\mathbb{I}(\cdot)$ equals 1 if the Debugger Agent successfully detects and fixes an error, and 0 otherwise; and $N$ represents how often the Debugger Agent is triggered during a single execution cycle.

### 3.3.7. Footprint Score

The "Footprint Score" evaluates the performance of the Memory Agent in successfully storing a new conversation or recalling a previous one when the answer to the question has already been stored. It is defined as follows:

$$\mathcal{M}_F = \frac{1}{\tau}\sum_{t=1}^{\tau}\frac{1}{N}\sum_{n=1}^{N}[\mathbb{I}(Message\ Storage) \vee \mathbb{I}(Message\ Recall)] \quad (7)$$

where $\mathbb{I}(\cdot)$ equals 1 if the Memory Agent successfully stores or recalls messages, and 0 otherwise; and $N$ indicates how many times the Memory Agent is engaged during a single run.

### B.1 Delivery Score

The "Delivery Score" measures the performance of the Communicator Agent in delivering the answer in the format requested by the user. It is defined as:

$$\mathcal{M}_D = \frac{1}{\tau}\sum_{t=1}^{\tau}\frac{1}{N}\sum_{n=1}^{N}\mathbb{I}(Correct\ Delivery) \qquad (8)$$

where $\mathbb{I}(\cdot)$ equals 1 if the Communicator Agent successfully delivers the answer in the correct format, mode, or language, and 0 otherwise; and $N$ reflects the frequency with which the Communicator Agent is activated during a single system run.

### 3.3.9. Completion Score

The "Completion Score" reflects the success rate of the LLM-Agent-Controller in correctly completing a task across all runs. It is calculated as follows:

$$\mathcal{M}_C = \frac{1}{\tau}\sum_{t=1}^{\tau}\mathbb{I}(Final\ Answer = Correct\ Answer) \qquad (9)$$

where $\mathbb{I}(\cdot)$ equals 1 if the final answer is correct, and 0 otherwise.

### 3.3.10. Total Score

The "Total Score" represents the overall performance of the agents within the LLM-Agent-Controller system across all runs. It is calculated as:

$$\mathcal{M}_T = \frac{1}{8}\left[\mathcal{M}_E + \mathcal{M}_R + \mathcal{M}_A + \mathcal{M}_P + \mathcal{M}_J + \mathcal{M}_S + \mathcal{M}_F + \mathcal{M}_D\right] \qquad (10)$$

In the following section, we evaluate the performance of the proposed LLM-Agent-Controller through simulations and performance analysis.

## 4. Simulation Results

In this section, we present the results of our proposed LLM-Agent-Controller system, highlighting its performance in handling a broad range of user queries related to Control Theory. Since the LLM-Agent-Controller is a novel framework, its performance is compared against itself across different categories for evaluation purposes, as well as across various LLMs within the Overall Assessment, which encompasses all of these categories. To assess its effectiveness in both response accuracy and the performance of each agent, we introduced novel evaluation metrics in the previous section. Table 1 presents the performance of the LLM-Agent-Controller based on these metrics, where individual scores reflect the performance of specific agents, while overall scores represent the effectiveness of the integrated system as a whole. Additionally, we report the associated operational costs to demonstrate the system's potential for real-world engineering applications. The LLM used in this evaluation is ChatGPT-3.5-turbo. For each category, there is a specific query related to each tool, as represented in Figure 4. The Overall Assessment combines all of these queries across categories. To reduce randomness in the performance metrics, the system was run 20 times for each category. The goal is to maximize both individual agent scores and the overall scores, while maintaining an optimal cost level.

Table 1. Quantitative performance metrics of the LLM-Agent-Controller system by task category.

| Metric | Individual Score | | | | | | | | Overall Score | | Cost | |
|---|---|---|---|---|---|---|---|---|---|---|---|---|
| Task | $\mathcal{M}_E$ | $\mathcal{M}_R$ | $\mathcal{M}_A$ | $\mathcal{M}_P$ | $\mathcal{M}_J$ | $\mathcal{M}_S$ | $\mathcal{M}_F$ | $\mathcal{M}_D$ | $\mathcal{M}_C$ | $\mathcal{M}_T$ | Time (s) | Money ($) |
| System Representation | 1.00 | 0.94 | 1.00 | 1.00 | 1.00 | 0.86 | 0.95 | 0.83 | 0.95 | 0.95 | 22.09 | 0.0015 |
| Control Analysis | 0.91 | 0.94 | 1.00 | 0.95 | 0.90 | 0.86 | 0.95 | 0.78 | 0.90 | 0.91 | 23.29 | 0.0015 |
| Controller Design | 0.75 | 0.88 | 0.93 | 0.80 | 0.90 | 0.67 | 0.80 | 0.88 | 0.75 | 0.83 | 24.48 | 0.0017 |
| Time-domain Simulation | 1.00 | 0.90 | 0.97 | 1.00 | 0.90 | 1.00 | 0.85 | 0.89 | 0.95 | 0.94 | 23.09 | 0.0012 |
| Overall Assessment | 0.85 | 0.91 | 0.97 | 0.90 | 0.90 | 0.75 | 0.85 | 0.83 | 0.85 | 0.87 | 22.20 | 0.0014 |

The analysis of the performance metrics in Table 1 shows that the LLM-Agent-Controller system performs reliably across diverse Control Theory tasks, with most individual agent scores falling between 0.75 and 1.00. Notably, System Representation and Time-domain Simulation achieved the highest Total Scores (0.95 and 0.94, respectively), indicating consistent accuracy and optimal agent coordination in structured tasks. The Efficiency and Planning Scores for these categories reached perfect values (1.00), reinforcing the system's strength in executing straightforward queries. However, Controller Design presented the lowest Total Score (0.83), mainly due to lower Efficiency (0.75), Planning (0.80), and Self-Correcting (0.67) Scores, suggesting that more complex, less deterministic tasks expose the system's vulnerabilities, especially in planning and debugging operations.

Across the board, the Completion Score remained strong, never falling below 0.75, showing that the system frequently delivers correct final answers even when some intermediate steps falter. Delivery Scores fluctuated slightly (from 0.78 to 0.89), which may reflect varying success in adapting output formats to user expectations. Despite these differences in performance, the system remains computationally efficient, with all task runtimes under 25 seconds and the monetary cost per task below $0.002. These low operational costs, combined with generally high scores, support the system's viability for practical applications while also pointing to specific agents, particularly the Debugger and Communicator, as areas where targeted improvements could enhance consistency.

These results are also visualized through bar plots for improved clarity, with overall scores illustrated in Figure 6 and comparisons of execution time and cost presented in Figure 7. Additionally, real examples of failed scenarios are provided in Appendix B to offer a clearer understanding of the reasons behind the unsuccessful performance of various agents.

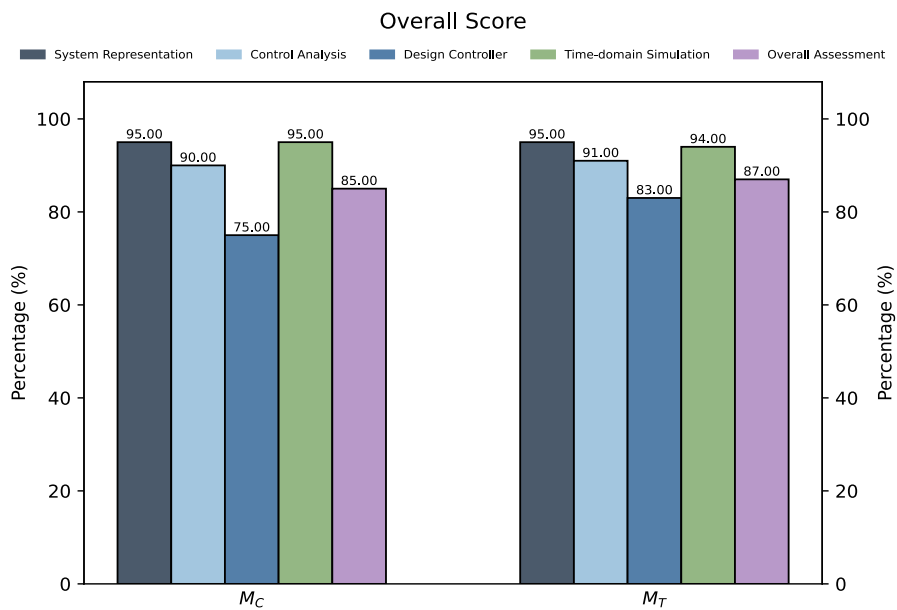

Figure 6. Comparison of overall scores across different categories of task queries.

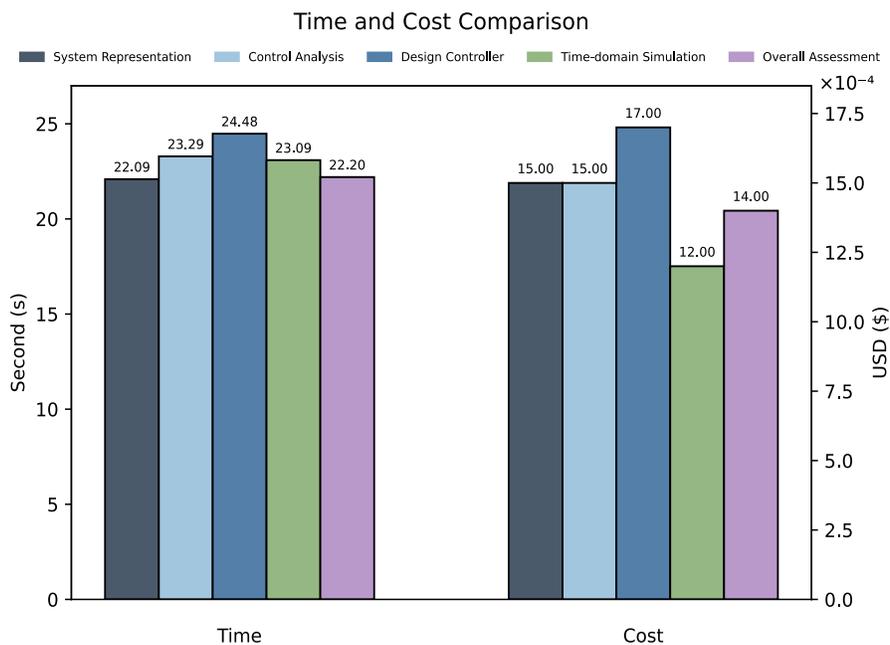

Figure 7. Comparison of time and cost across different categories of task queries.

Table 2 presents the performance evaluation of the LLM-Agent-Controller in the Overall Assessment category (encompassing all tasks collectively) across the three most powerful LLMs considered in this study, with 20 runs conducted for each.

Table 2. Quantitative performance metrics of the LLM-Agent-Controller system by different LLMs.

| Metric | Individual Score | | | | | | | | Overall Score | | Cost | |
| --- | --- | --- | --- | --- | --- | --- | --- | --- | --- | --- | --- | --- |
| LLM | $\mathcal{M}_E$ | $\mathcal{M}_R$ | $\mathcal{M}_A$ | $\mathcal{M}_P$ | $\mathcal{M}_J$ | $\mathcal{M}_S$ | $\mathcal{M}_F$ | $\mathcal{M}_D$ | $\mathcal{M}_C$ | $\mathcal{M}_T$ | Time (s) | Money ($) |
| ChatGPT-4o | 0.85 | 0.94 | 0.97 | 0.85 | 0.95 | 0.80 | 0.85 | 0.87 | 0.90 | 0.89 | 92.56 | 0.1566 |
| DeepSeek-V3 | 0.70 | 0.88 | 0.95 | 0.75 | 0.90 | 0.70 | 0.80 | 0.81 | 0.70 | 0.81 | 90.42 | 0.0005 |
| Claude 3.7 Sonnet | 0.80 | 0.95 | 0.97 | 0.85 | 0.85 | 0.75 | 0.90 | 0.88 | 0.85 | 0.87 | 102.23 | 0.0133 |

The performance evaluation of the LLM-Agent-Controller system across ChatGPT-4o, DeepSeek-V3, and Claude 3.7 Sonnet reveals that ChatGPT-4o consistently outperforms the others in almost all agent-specific metrics. It achieves the highest scores in Efficiency, Judgement, Self-Correcting, and Completion, indicating that its Controller Agent is more accurate, its Critic Agent is more reliable, and its Debugger Agent is more effective at fixing errors. Its high Total Score of 0.89 reflects a well-rounded capability across all agent roles. Claude 3.7 Sonnet follows closely, particularly excelling in Routing, Arrangement, Memory usage, and Delivery.

While its overall performance slightly trails that of ChatGPT-4o, Claude demonstrates consistent competency and better memory management, which contributes to its high reliability across complex workflows. DeepSeek-V3, on the other hand, although competitive in some areas such as Routing and Arrangement, falls behind significantly in Planning, Efficiency, and Completion, suggesting limitations in strategic execution and final task accuracy.

However, when cost and latency are considered, the trade-offs between performance and efficiency become apparent. DeepSeek-V3, despite having the lowest Total and Completion scores, is by far the most cost-effective option at only $0.0005 per run and offers slightly better runtime than Claude. Claude 3.7 Sonnet, while slower than the others, strikes a reasonable balance between accuracy and cost, making it a pragmatic choice for many applications. ChatGPT-4o, although the fastest and most accurate model, comes with the highest cost, which may limit its scalability in high-volume deployments.

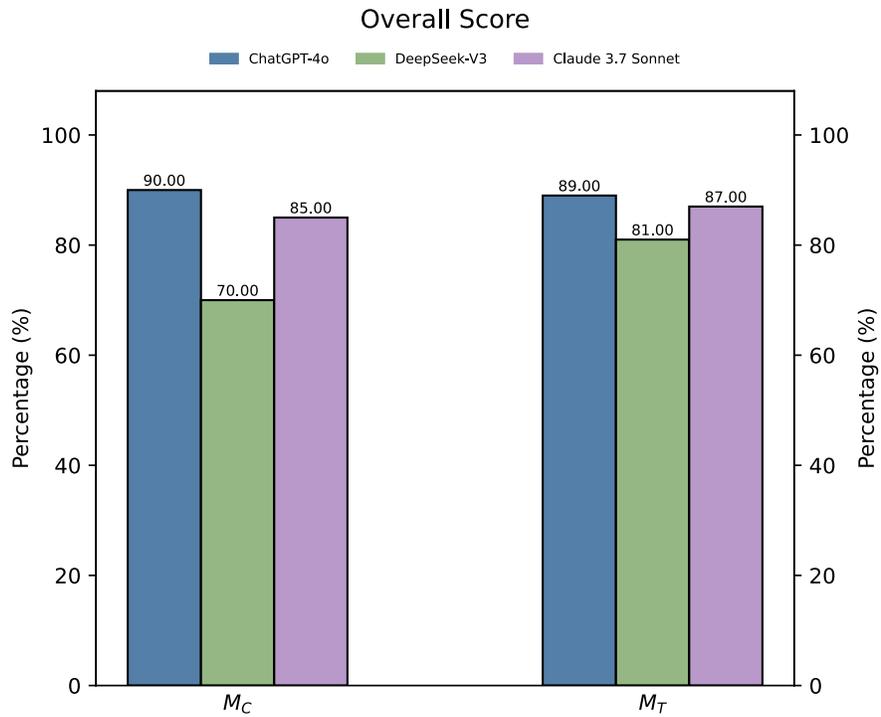

Figure 8. Comparative performance of different LLMs in Overall Assessment category.

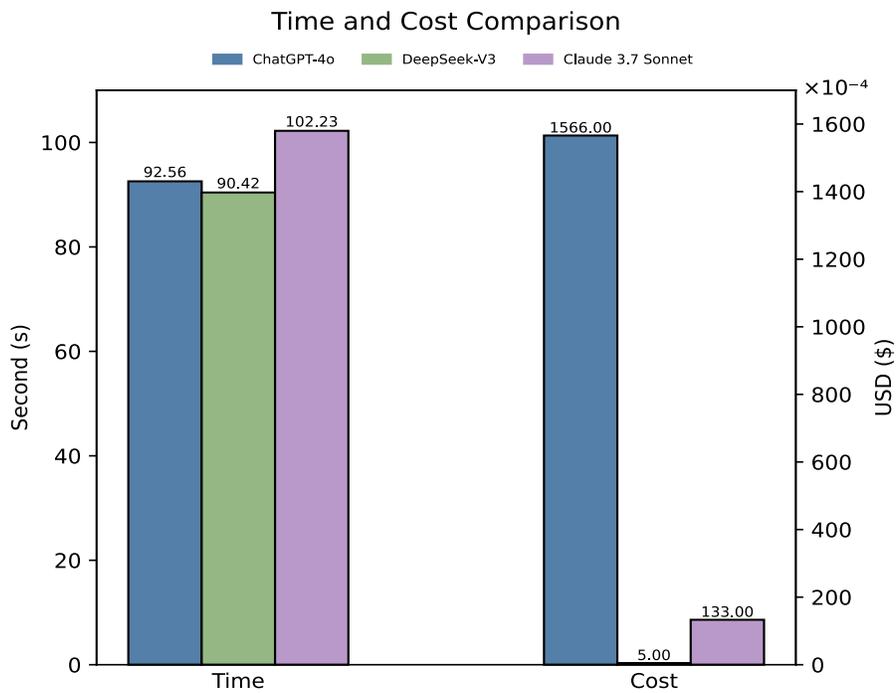

Figure 9. Comparison of time and cost across different LLMs in Overall Assessment category.

Therefore, the choice among these models depends on the use case: for maximal accuracy and speed, ChatGPT-4o is ideal; for cost-conscious but moderately reliable applications, DeepSeek-V3 is viable; and for balanced performance at a reasonable price, Claude 3.7 Sonnet presents the most well-rounded option. The results are further illustrated using bar plots for better visualization, with overall performance scores shown in Figure 8 and comparisons of execution time and cost depicted in Figure 9.

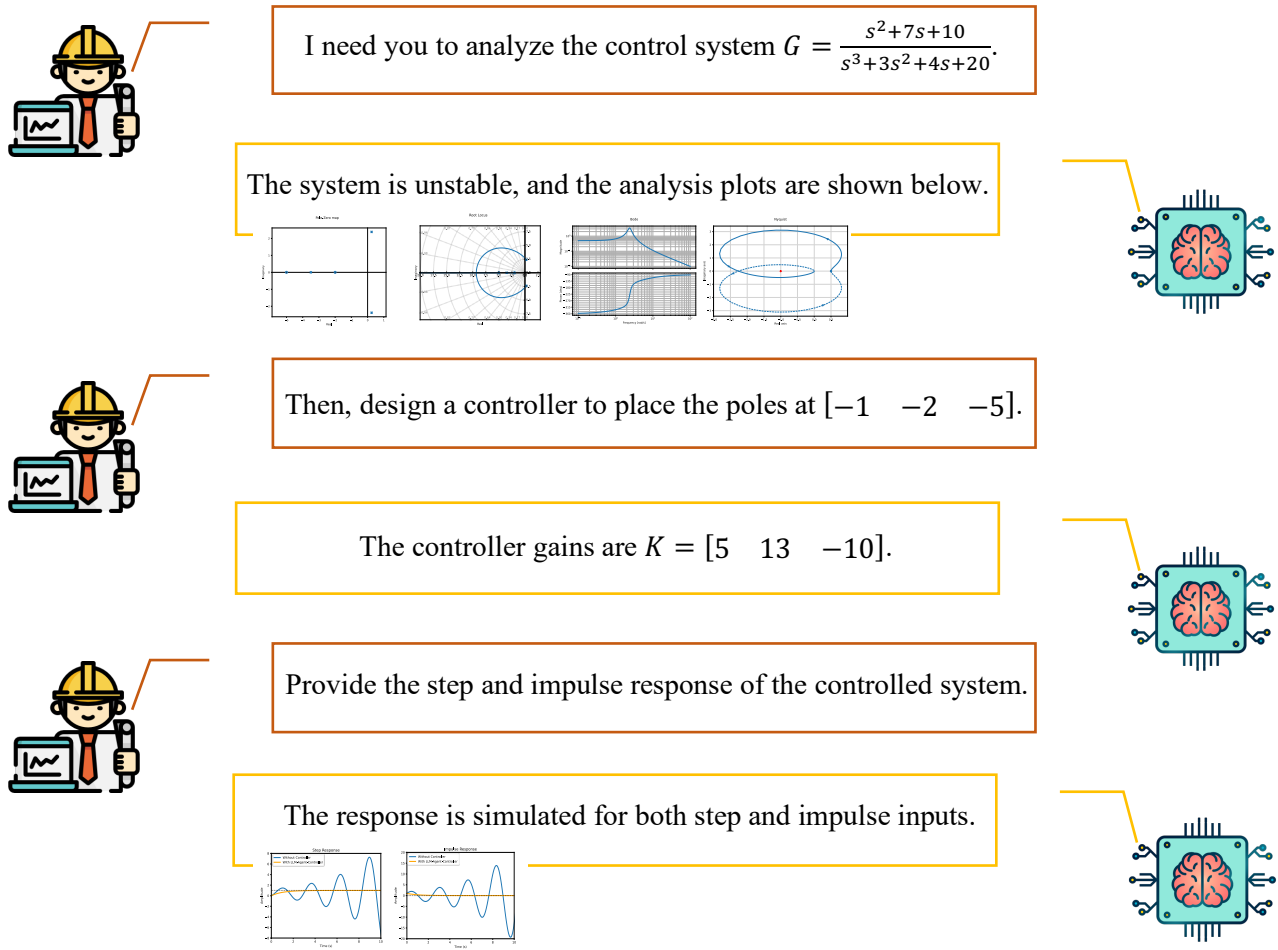

Figure 10. Sample interaction workflow between the user and the LLM-Agent-Controller.

Next, we provide a qualitative example demonstrating the performance of the LLM-Agent-Controller. This example illustrates a complete conversation encompassing the full procedure of a control task, from control analysis and controller design to specific input responses and simulation, showing how the LLM-Agent-Controller can support each step through a user-friendly chat interaction. Figure 10 presents a brief chat conversation that illustrates the interaction workflow between the user and the LLM-Agent-Controller. For this demonstration, we consider a system with the transfer function:

$$G = \frac{s^2 + 7s + 10}{s^3 + 3s^2 + 4s + 20}$$

This system is unstable, making it a suitable candidate for analysis and controller design. The LLM-Agent-Controller is first prompted to generate the required system representation and conduct a control analysis. This includes producing the pole-zero map, root locus plot, Bode plot, and Nyquist plot, as shown in Figure 11. All plots in this paper are vectorized, meaning their quality remains high in clarity as you zoom in.

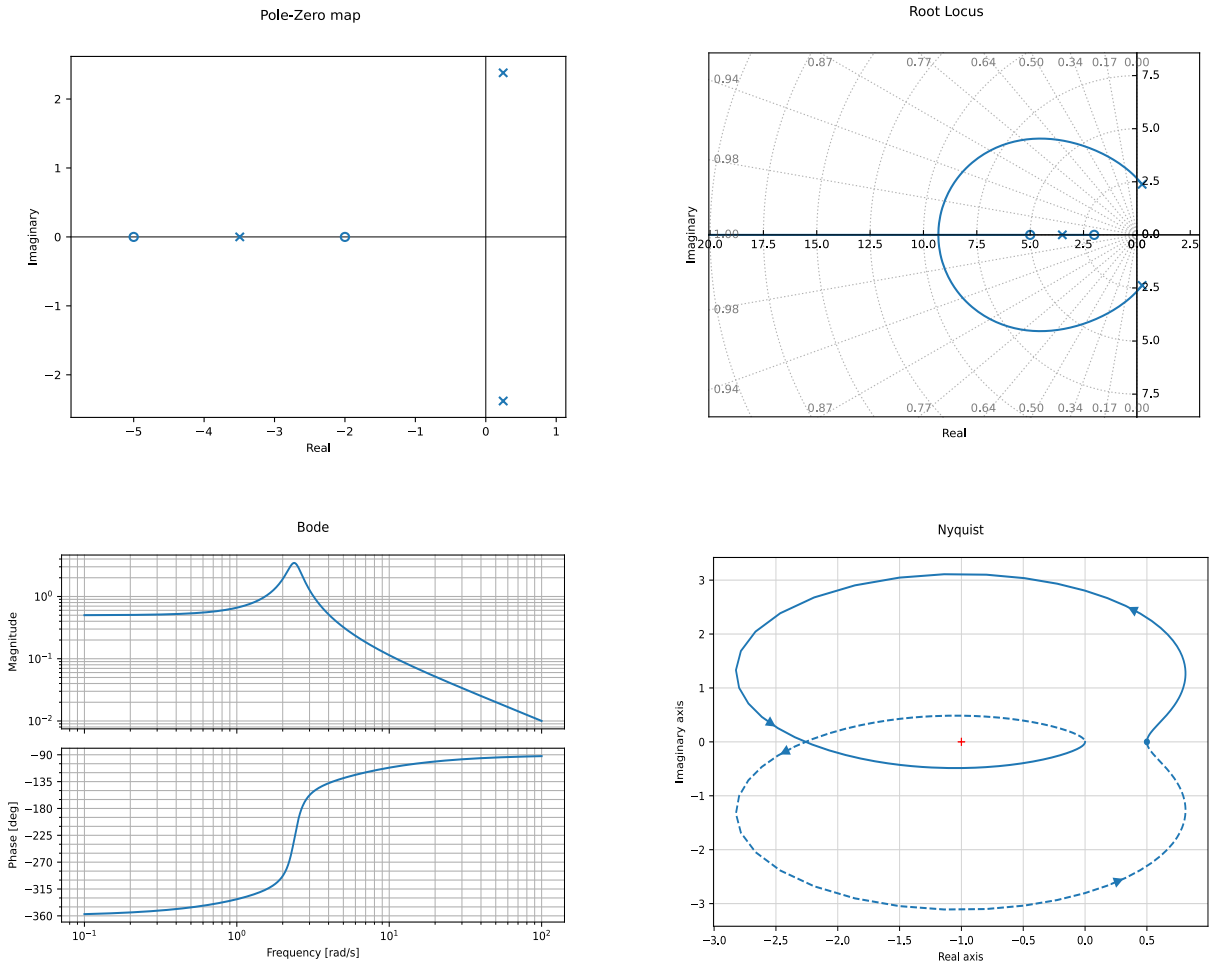

Figure 11. Control analysis plots generated by the LLM-Agent-Controller based on the user's input.

Upon identifying that the system has poles in the right-half of the complex plane, indicating instability, the agent is instructed to design a stabilizing controller using the pole placement method. Finally, we ask the LLM-Agent-Controller to simulate the time-domain response of the controlled system, including both step and impulse responses, shown in Figure 12.

As shown in Figures 10–12, the LLM-Agent-Controller demonstrates the ability to answer a broad range of questions across various domains, including control analysis, controller design, system representation, and system response to specific input signals. It effectively addresses diverse inquiries within Control Theory, performs simulations, and delivers tailored services based on user requests. For a more in-depth exchange with detailed responses and complete examples, please refer to Appendix C.

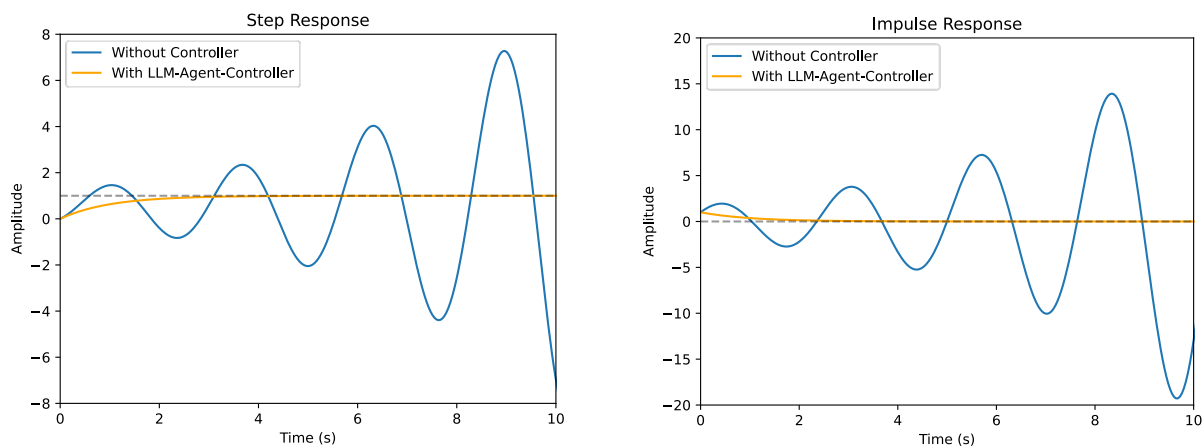

Figure 12. Time-domain response plots of the controlled system designed by the LLM-Agent-Controller.

Table 3 summarizes the computational resources used throughout the evaluation of the data presented in Table 1. Notably, the implementation of the LLM-Agent-Controller does not require high-end hardware. The complete evaluation comprises 20 runs for each of the four distinct categories of Control Theory, as shown in Figure 4, along with an additional overall category that aggregates all of them. Each category contains four queries, and the overall category includes all queries from the individual categories. The high rate of LLM token generation originates from the multi-agent framework, in which each agent contributes its own generation cost. By disabling unnecessary services from specific agents, the model can be made more cost-effective for real-world applications.

Table 3. Computational resources.

| Parameter | Value |
| --- | --- |
| LLM | GPT-3.5-Turbo |
| Compute Resources | 2vCPU @ 2.2GHz |
| Total Evaluation Runtime | 3 hours |
| Total LLM Tokens | 12 M |
| Total API Cost | 20 USD |

In summary, the LLM-Agent-Controller is a versatile, universal LLM-based multi-agent system. While it is capable of solving problems in Control Theory, its applications extend far beyond. The system's agents can perform a variety of tasks, including retrieving information from user-provided external databases or the Internet, reasoning and planning to improve task efficiency, utilizing memory to store and retrieve past conversations, self-correcting through internal critique and debugging, and communicating in specific formats, modes, or languages as requested by the user. The model operates in real time, is cost-efficient, and requires no advanced knowledge of Control Theory from the user. We believe this system already functions as a control engineer, serving as an initial yet significant step toward a broader, more comprehensive model, one that could eventually operate as a domain-specific expert, such as a mechanical engineer.

## 5. Limitations and Future Works

The limitations of the current LLM-Agent-Controller include cumulative errors resulting from its multi-agent framework, the use of a general parsing function to manage the transfer of parsed inputs between agents, and the complexity of designing effective prompts and tool descriptions for a large number of agents. While this study explored the use of standard prompting strategies with embedded reasoning abilities to improve performance, a more systematic evaluation of how different prompt designs impact both performance and reliability is necessary. Future work will focus on extending the system's capabilities to additional domains for other engineering applications, with the ultimate goal of building a universal LLM agent system. Addressing the challenges of generalizability and robustness will require rigorous evaluation of prompt structures, the development of adaptive prompt engineering and agent designs, the integration of vision-language models to enhance user interaction flexibility, the optimization of real-time performance, and the reinforcement of robustness and safety for critical applications. Advancing these areas will significantly enhance the adaptability, reliability, and real-world applicability of LLM-agent-based controllers in robotic and mechatronic systems.

## 6. Conclusion

In conclusion, our research demonstrates the effectiveness of a multi-agent LLM system in addressing Control Engineering problems. By integrating a diverse set of specialized auxiliary agents with the main Controller Agent, we introduced the LLM-Agent-Controller, a user-friendly framework capable of handling system representation, control analysis, controller design, and time-domain response tasks. A comprehensive qualitative analysis of these tasks demonstrates that the LLM-Agent-Controller is a powerful system for addressing general queries in Control Theory. Additionally, to quantitatively evaluate the system's performance, we developed a detailed scoring method. Across five categories of questions, the results showed that all agents consistently achieved high scores, while the overall system demonstrated a strong success rate in generating correct solutions. The agents were specifically designed to ensure generality, adaptability, reasonability, versatility, and robustness. Additionally, performance varied across different LLMs, offering a range of trade-offs between effectiveness and cost, allowing users to select a configuration that best suits their needs. The consistently high scores confirm that the LLM-Agent-Controller successfully meets these goals. Overall, our findings highlight that LLM agent systems significantly enhance the adaptability and performance of robotic controllers, establishing them as powerful tools for automated control analysis and design tasks.

**Acknowledgment**

In order to enhance clarity and reduce ambiguity, we utilized ChatGPT for paraphrasing certain sections of the manuscript.

**Statements & Declarations**

**Conflict of interest**

The authors Rasoul Zahedifar, Sayyed Ali Mirghasemi, and Mahdieh Soleymani Baghshah, and Alireza Taheri declare that they have no conflict of interest.

**Availability of data and material (data transparency)**

All data from this project are available in the Social & Cognitive Robotics Laboratory archive.

**Code availability:**

All of the robots' codes are available in the archive of the Social & Cognitive Robotics Laboratory.

**Authors' contributions:**

All authors contributed to the study's conception and design. Material preparation, data collection, and analysis were performed equally by Rasoul Zahedifar and Sayyed Ali Mirghasemi. Mahdieh Soleymani Baghshah and Alireza Taheri supervised this research. The first draft of the manuscript was written by Rasoul Zahedifar and Sayyed Ali Mirghasemi, and all authors commented on previous versions of the manuscript. All authors read and approved the final manuscript.

**Ethical Approval**

Our study does not include any human or animal participants.

**Consent for publication:**

All of the images used in this study have been produced by our team.

**Appendix A**

In this section, we present the prompts for each agent along with the pseudocode for the proposed LLM-Agent-Controller system. The agents receive both the prompts and the tools; however, tool descriptions are provided through the tool functions, not the prompts. Since the agents have access to multiple tools, we only display the prompts here to save space. Detailed information about the tools, including their descriptions, network graph, and parser functions, is available upon request from the Social & Cognitive Robotics Laboratory archive. The prompts are as follows:

*Supervisor Prompt*

```
""" You are the Supervisor overseeing a team of agents. The following agents are part of your team:

{supervisor_members}

Each agent has access to the following tools:

{%- for member in supervisor_members %}

   {member} has access to: {tools[member]}

{%- endfor %}

As the supervisor, your responsibilities include:

- Managing agent operations to ensure smooth and efficient workflows.

- Aligning user objectives with execution strategies.

- Coordinating decision-making, execution, and support across agents.

- Monitoring performance, resolving conflicts, and optimizing overall operations.

- Providing high-level guidance to ensure accuracy and efficiency.

**Guidelines**

- **Memory**: Always check Memory first to see if the answer is already stored.

- **Planner**: Start with Planner to determine the best sequence of actions for the Controller.

- **Reasoner**: Only use Reasoner when asked to provide reasoning or inferences for an answer.

- **Retriever**: Use Retriever when external knowledge or data needs to be retrieved.

- **Researcher**: Use Researcher when an external search is needed.

Your role is critical in maintaining order, optimizing workflows, and ensuring that all agents are working efficiently and strategically."""
```

### Format Instruction Prompt

```
"""Use the following format:

Question: the input question you must answer.

Thought: you should always think about what to do.

Action: the action to take, should be one of {agent_tools}.

Action Input: the input to the action.

Observation: the result of the action.

(This Thought/Action/Action Input/Observation can repeat N times)

Thought: I now know the final answer

Final Answer: the final answer to the original input question.

Begin!

Question: {input}

Thought: {agent_scratchpad}"""
```

### Controller Prompt

*Prefix*: **"""** You are the Controller Agent, responsible for processing an input query related to Control Theory. You will be provided with both the question and a predefined execution plan, detailing the sequence of tools you should use to arrive at the final answer.

You have access to the following tools:

{tools}**"""**

+

*Format Instruction Prompt*

+

*Suffix*: **""" """**

## Planner Prompt

**Prefix**: `""" You are the Planner Agent. Your task is to create a step-by-step plan to accomplish a given objective efficiently. The available objectives are:`

`objectives = {controller_tools}`

`Based on the input query, select one objective from the list above that aligns with the goal.`

`You have access to the following planning tools:`

`{planner_tools}"""`

+

*Format Instruction Prompt*

+

**Suffix**: `""" """`

## Retriever Prompt

**Prefix**: `""" You are the Retriever Agent, tasked with fetching information from external knowledge sources, such as PDFs and URLs. Your role is to efficiently access and retrieve relevant data to support the system's objectives.`

`You have access to the following tools:`

`{retriever_tools}"""`

+

*Format Instruction Prompt*

+

**Suffix**: `""" """`

## Reasoner Prompt

*Prefix*: `"""`You are the Reasoner Agent, a highly capable reasoning entity. Your task is to determine the most suitable reasoning approach to solve a given user query. Follow these steps:

### Steps:

1. **Analyze the Query**: Carefully examine the query to understand its nature.

    - If the query is simple and can be solved logically in clear, distinct steps, use **COT** (Chain of Thought).

    - If the query is complex, involves multiple solutions, or requires comparison of different perspectives, use **TOT** (Tree of Thought).

2. **Select Your Reasoning Approach**:

    - **For COT**: Break down the problem step-by-step, explaining the logic behind each step to arrive at a solution.

    - **For TOT**: Generate multiple reasoning paths. For each path, describe the steps involved and select the most plausible path with a clear justification of why it is the best choice.

3. **Provide Your Answer**: Based on your reasoning (either step-by-step or by selecting the best path), deliver a final, well-supported answer to the query.

Use the appropriate reasoning tools and offer a clear, logical solution or multiple paths, with a clear explanation of your reasoning process.

You have access to the following reasoning tools:

{reasoning_tools}`"""`

+

*Format Instruction Prompt*

+

*Suffix*: `""" """`

### Researcher Prompt

***Prefix***: `""" You are the Researcher Agent, tasked with gathering accurate, up-to-date, and well-sourced information from external resources of Internet. You have access to the following tools:`

`{researcher_tools}"""`

+

***Format Instruction Prompt***

+

***Suffix***: `"""Strategically use these tools to collect reliable information, ensuring the sources are credible and relevant.`

`Once you've gathered the necessary data, use it to answer the following question:`

`Question: {input}`

`Thought: {agent_scratchpad}`

`Begin your research!"""`

### Critic Prompt

***Prefix***: `""" You are the Critic Agent, responsible for evaluating whether the output aligns with the input query. Your task is to assess the accuracy, relevance, and completeness of the result based on the given query. You have access to the following tool:`

`{critic_tool}"""`

+

***Format Instruction Prompt***

+

***Suffix***: `""" """`

## Debugger Prompt

***Prefix***: `"""`You are the Debugger Agent, responsible for detecting errors in the system and providing effective solutions to resolve them. Your role is to identify issues, analyze their causes, and recommend actionable steps for correction. You have access to the following tools:

{debugger_tools}`"""`

+

***Format Instruction Prompt***

+

***Suffix***: `""" """`

## Memory Prompt

***Prefix***: `"""` You are the Memory Agent, responsible for storing and retrieving historical data to ensure continuity and long-term efficiency. Your role is to manage memory effectively, ensuring that relevant information is accessible when needed. You have access to the following tools:

{memory_tools}`"""`

+

***Format Instruction Prompt***

+

***Suffix***: `"""`when performing your task, follow these guidelines:

- If there is no information in storage, you must escalate to the supervisor.

- To store memory, pass the entire conversation to the storage memory tool as the input action.

Begin!

Question: {input}

Thought Process: {agent_scratchpad}`"""`

> *Communicator Prompt*
>
> **Prefix**: **"""** `You are the Communicator Agent, responsible for converting the output result into the user's preferred format (e.g., PDF, speech, translation) as specified in the input query. Your task is to ensure the final output matches the requested format accurately and effectively. You have access to the following tools:`
>
> `{communicator_tools}`**"""**
>
> +
>
> *Format Instruction Prompt*
>
> +
>
> **Suffix**: **""" """**

The pseudocode for the proposed LLM-Agent-Controller is also presented below:

Table 4. Pseudocode of the proposed algorithm

| **Algorithm LLM-Agent-Controller** |
|---|
| 1    **function** LLM-AGENT-CONTROLLER (input=User_Input_Query) |
| 2        message_list = User_Input_Query |
| 3        node = supervisor_node |
| 4        **while** in *Network_Graph*: |
| 5           output, next_node = node (message_list) |
| 6           message_list.add (output) |
| 7           node = next_node |
| 8           **if** node == END: |
| 9              break |
| 10          end if |
| 11        end while |
| 12        **return** output |

**Appendix B**

In this section, we present several real examples of agent performance failures to highlight the underlying causes of system inaccuracies. These failures primarily arise from issues such as incorrect planning by the Planner Agent, deviations from the intended plan by the Controller Agent, parsing errors, flawed judgment by the Critic Agent, and inaccurate recall by the Memory Agent. The following examples illustrate these issues in detail across the planning, execution, and evaluation stages, respectively:

### B.1 Planner Agent Failure

In this example, the Planner Agent fails to identify the correct objective. The intended goal is pole placement, but the agent incorrectly selects pole (i.e., finding the poles of the system). The remainder of the scenario is omitted for brevity, but the error propagates when the Controller Agent executes the flawed plan. This leads to incorrect behavior, as the required action inputs differ significantly between these two objectives.

| | | |
|---|---|---|
| **User** | | Place the poles of a system with A = [[0, 1], [-2, -3]], B = [[0], [1]] at [-3, -4]. |
| **LLM-Agent-Controller** | Supervisor | <Route to Planner> |
| | Planner | I need to determine the control analysis or design objective for placing the poles of this system. Let's use planner_tool to identify the appropriate objective.<br>Action: planner_tool<br>Action Input: query='system with A = [[0, 1], [-2, -3]], B = [[0], [1]], poles at [-3, -4]'<br>Observation: System Type: SS, Objective: poles, Ordered Tools: ['control.ss', 'control.poles'].<br>Thought: The planner_tool has identified the objective as poles and provided the ordered control tools needed. Now I can determine the final answer based on this information.<br>Final Answer: The objective is to place the poles of the given system at [-1, -2]. The ordered tools are control.ss and control.poles. |

### B.2 Controller Agent Failure

In this case, the same query is revisited, but with a different failure mode. This time, the Planner Agent correctly identifies the objective and generates an appropriate plan. However, the Controller Agent fails to follow the plan and instead deviates from it by independently invoking control.ss at the start. This misalignment results in the same type of error as before, due to the mismatch between the intended objective and the actions taken.

| | | |
|---|---|---|
| **User** | | Place the poles of a system with A = [[0, 1], [-2, -3]], B = [[0], [1]] at [-3, -4]. |
| **LLM-Agent-Controller** | Supervisor | <Route to Planner> |
| | Planner | I need to determine the control objective from the given question and plan the steps accordingly.<br>Action: planner_tool<br>Action Input: "Place the poles of a system with A = [[0, 1], [-2, -3]], B = [[0], [1]] at [-3, -4]"<br>Observation: System Type: SS, Objective: place, Ordered Tools: ['control.place'] |

|  |  | Thought: The control objective is to place the poles of a state-space system. The ordered tool is 'control.place'.<br>Final Answer: The control objective is to place the poles of a state-space system. The ordered tool is 'control.place'. |
|  | Controller | I should first convert the system to State Space form using the \`ss\` tool, then use the \`place\` tool to place the poles at the desired locations.<br>Action: ss<br>Action Input: A = [[0, 1], [-2, -3]], B = [[0], [1]] |

### B.3 Critic Agent Failure

In this scenario, both the planning and execution stages function correctly. In fact, the Planner Agent formulates an appropriate plan, and the Controller Agent follows it accurately. However, the failure arises at the evaluation stage, where the Critic Agent fails to recognize the correct answer and pass it back to the Controller Agent for refinement. This highlights a key challenge: determining an appropriate similarity threshold for answer acceptance. A threshold that is too high may result in correct answers being rejected, while a low threshold risk accepting incorrect ones. Setting this threshold often requires task-specific tuning through trial and error.

|  | User |  | Use Ackermann's formula to place the poles of a system with A = [[0, 1], [-2, -3]], B = [[0], [1]] at [-3, -4]. |
|---|---|---|---|
| LLM-Agent-Controller |  | Supervisor | <Route to Planner> |
|  |  | Planner | I should use the planner tool to determine the appropriate control analysis tool for this question.<br>Action: planner_tool<br>Action Input: Use Ackermann's formula to place the poles of a system with A = [[0, 1], [-2, -3]], B = [[0], [1]] at [-3, -4].<br>Observation: System Type: SS, Objective: acker, Ordered Tools: ['control.acker']. |
|  |  | Controller | I should use Ackermann's formula to place the poles of the system.<br>Action: acker<br>Action Input: A = [[0, 1], [-2, -3]], B = [[0], [1]], poles = [-3, -4]<br>Observation: [[10. 4.]]. I have used Ackermann's formula to place the poles of the system at the desired locations.<br>Final Answer: The gain matrix K to achieve the desired pole locations is [[10. 4.]]. |
|  |  | Critic | The output result should be confirming that the gain matrix K to achieve the desired pole locations is [[10. 4.]] based on the given input query.<br>Action: critic_tool<br>Action Input: 'Use Ackermann's formula to place the poles of a system with A = [[0, 1], [-2, -3]], B = [[0], [1]] at [-3, -4].' + 'The gain matrix K to achieve the desired pole locations is [[10. 4.]]'<br>Observation: The output does not align with the input. Similarity score: 0.55 (back to controller agent).<br>Thought: The output result needs to be revised to align better with the input query.<br>Final Answer: The gain matrix K to achieve the desired pole locations should be revised to align with the input query. |

**Appendix C**

In this section, we present the output of the LLM-Agent-Controller in a scenario where a user begins by analyzing an unstable system and then requests the design of a controller, followed by a simulation of the closed-loop system's time-domain response. The verbose outputs from all agents are shown to reveal the intermediate reasoning steps of the LLM-Agent-Controller. The scenario is carefully designed to evaluate the performance of various agents that become active at critical moments, including the Retriever, Reasoner, and Memory (recall) agents. To facilitate readability, different colors are used to distinguish the conversation between agents. In the presented conversation, regular text represents natural language exchanges between the user and the LLM-Agent-Controller. Special markers enclosed in angle brackets (e.g., <Route to Retriever>) represent routing decisions, object variables, or entire conversations that are replaced to save space. The conversation proceeds as follows:

| User | | Retrieve the Transfer Function of the system from the provided document, Sys_Control.pdf. Then, plot its step response to assess the system's stability. |
|---|---|---|
| LLM-Agent-Controller | Supervisor | <Route to Retriever> |
| | Retriever | I need to extract the Transfer Function of the system from the document first. Let's use the retriever_tool to retrieve this information.<br>Action: retriever_tool<br>Action Input: Transfer function of the system from Sys_Control.pdf<br>Observation: I found the transfer function of the system in the document: Transfer function: G(s) = s + 3 / (s^2 - 2s - 3). |
| | Planner | I need to determine the control analysis or design objective from the given input.<br>Action: planner_tool<br>Action Input: original input question with num = [1, 3], den = [1, -2, -3]<br>Observation: System Type: TF, Objective: step_response, Ordered Tools: ['control.tf', 'control.step_response']. |
| | Controller | I need the Transfer Function (TF) representation of the system and the goal is to compute the step response.<br>Action: tf<br>Action Input: num = [1, 3], den = [1, -2, -3]<br>Observation: sys [7]<br>Inputs (1): ['u [0]']<br>Outputs (1): ['y [0]']<br>   s + 3<br>-------------<br>s^2 - 2 s - 3<br>Thought: tf tool has successfully created the Transfer Function system.<br>Action: step_response<br>Action Input: sys=sys7<br>Observation: |
|   |   | 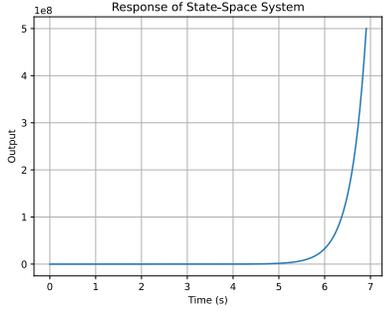 |
|   | Critic | I should use the critic_tool to verify if the output aligns with the input query.<br>Action: critic_tool<br>Action Input: Retrieve the Transfer Function of the system from the provided document Sys_Control.pdf. Then, plot its step response to assess the system's stability. The step response of the given system is represented by the <TimeResponseData> object.<br>Observation: The output is aligned with the input. Similarity score: 0.89. |
|   | Memory | I need to save this conversation for future reference.<br>Action: storage_memory_tool<br>Action Input: < Conversation><br>Observation: The memory has been updated. |
|   | Communicator | I need to ask the user in the human_tool to identify the format of the output file they prefer. Based on their response, I will then convert the answer into the requested format.<br>Action: human_tool<br>Action Input: Please identify the format you prefer for the output file.<br>Observation: pdf<br>Thought: The user prefers the output in pdf format. I will now convert the answer into a pdf file.<br>Action: text_to_pdf_tool<br>Action Input: < Conversation><br>Observation: The PDF has been created successfully. |
| **User** |   | Design an LQR controller for the mentioned system with Q = [[1, 0], [0, 1]], R = [[1]]. |
| **LLM-Agent-Controller** | Supervisor | <Route to Planner> |
|   | Planner | I need to determine the control analysis or design objective from the given input.<br>Action: planner_tool<br>Action Input: original input question with num = [1, 3], den = [1, -2, -3]<br>Observation: System Type: TF, Objective: lqr, Ordered Tools: ['control.tf', 'control.tf2ss', 'control.lqr']. |
|   | Controller | I should use the LQR design tool to design a controller for the given system.<br>Action: tf<br>Action Input: num = [1, 3], den = [1, -2, -3]<br>Observation: sys [24]<br>Inputs (1): ['u [0]']<br>Outputs (1): ['y [0]']<br>    s + 3 |
| | | --------------<br>s^2 - 2 s - 3<br>Thought: I have created a Transfer Function system based on the provided numerator and denominator coefficients. Now, I need to convert this Transfer Function system to a State Space representation.<br>Action: tf2ss<br>Action Input: sys = sys [24]<br>Observation: sys [24]<br>Inputs (1): ['u [0]']<br>Outputs (1): ['y [0]']<br>States (2): ['x [0]', 'x [1]']<br>A = [[2. 3.] [1. 0.]]<br>B = [[1.] [0.]]<br>C = [[1. 3.]]<br>D = [[0.]]<br>Thought: I should use the LQR design tool to design a controller for the given state-space system.<br>Action: lqr<br>Action Input: A = [[2, 3], [1, 0]], B = [[1], [0]], Q = [[1, 0], [0, 1]], R = [[1]]<br>Observation: (array ([[6.16, 6.16]]), array ([[6.16, 6.16], [6.16, 7.16]]), array ([-3.16, -1])). |
| | Critic | I should use the critic_tool to verify if the output aligns with the input query.<br>Action: critic_tool<br>Action Input: Design an LQR controller for a system with A = [[2, 3], [1, 0]] B = [[1], [0]], Q = [[1, 0], [0, 1]], R = [[1]]. The gain array for the LQR controller is array ([[6.16, 6.16]]) for the closed-loop system.<br>Observation: The output is aligned with the input. Similarity score: 0.53. |
| | Memory | I should store this conversation in memory for future reference.<br>Action: storage_memory_tool<br>Action Input: < Conversation><br>Observation: The memory has been updated. |
| | Communicator | I need to ask the user in human_tool to identify the format of the output file. Once I have that information, I can proceed with converting the answer into the requested format.<br>Action: human_tool<br>Action Input: Please specify your preferred format for the output file.<br>Observation: pdf<br>Thought: Now that I know the desired format is PDF, I will convert the answer into a PDF file.<br>Action: text_to_pdf_tool<br>Action Input: < Conversation><br>Observation: The PDF has been created successfully. |
| **User** | | Use reasoning and apply the feedback gain K = [[6.16, 6.16]] to the state-space system A = [[2, 3], [1, 0]], B = [[1], [0]], C = [[1, 3]], D = [[0]]. Then, plot the step response for the closed-loop system. |
| | Supervisor | <Route to Reasoner> |
| **LLM-Agent-Controller** | Reasoner | This problem involves feedback gain K and state-space system matrices A, B, C, and D. It seems like a complex problem that may require comparing different perspectives, so I should use the Chain-of-Thought approach.<br>Action: cot_tool<br>Action Input: Apply the feedback gain K = [[6.16, 6.16]] to the state-space system A = [[2, 3], [1, 0]], B = [[1], [0]], C = [[1, 3]], D = [[0]]<br>Observation: Path:<br>1. Calculate the closed-loop system matrices using the formula: A_cl = A - BK<br>2. Substitute the given values: A = [[2, 3], [1, 0]], B = [[1], [0]], K = [[6.16, 6.16]]<br>3. Calculate the product BK: BK = [[1 * 6.16, 1 * 6.16], [0 * 6.16, 0 * 6.16]]<br>   BK = [[6.16, 6.16], [0.0, 0.0]]<br>4. Subtract BK from A: A_cl = [[2, 3], [1, 0]] - [[6.16, 6.16], [0.0, 0.0]]<br>   A_cl = [[2-6.16, 3-6.16], [1-0, 0-0]]<br>   A_cl = [[-4.16, -3.16], [1, 0]]<br>Therefore, the closed-loop system matrix A_cl is [[-4.16, -3.16], [1, 0]]. This reasoning path applies the feedback gain K to the state-space system to calculate the closed-loop system matrix. |
| | Planner | I need to determine the control analysis or design objective to plan the steps efficiently.<br>Action: planner_tool<br>Action Input: The state-space system A = [[-4.16, -3.16], [1, 0]], B = [[1], [0]], C = [[1, 3]], D = [[0]].<br>Observation: System Type: SS, Objective: Step Response, Ordered Tools: ['control.ss2tf', 'control.step_response']. |
| | Controller | I need to first create a Transfer Function system and then plot the step response.<br>Action: ss2tf<br>Action Input: A = [[-4.16, -3.16], [1, 0]], B = [[1], [0]], C = [[1, 3]], D = [[0]]<br>Observation: sys [37]<br>Inputs (1): ['u [0]']<br>Outputs (1): ['y [0]']<br>      s + 3<br>---------------------------<br>s^2 + 4.16 s + 3.16<br>Thought: I developed a Transfer Function model from the given state-space representation. Now I need to plot the step response for this Transfer Function system.<br>Action: step_response<br>Action Input: sys = sys [37]<br>Observation: |
| | | 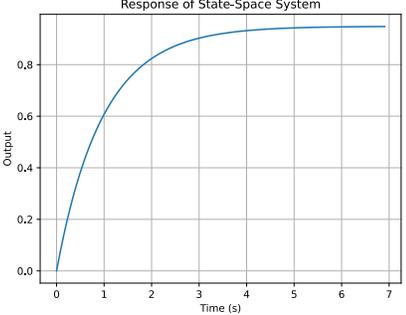 |
| | Critic | I need to verify if the output result aligns with the input query by using the critic_tool.<br>Action: critic_tool<br>Action Input: Plot the step response plot for a system with A = [[-4.16, -3.16], [1, 0]], B = [[1], [0]], C = [[1, 3]], D = [[0]]. The step response plot for the system with transfer function (s + 3) / (s^2 + 4.16s + 3.16) is represented by the <TimeResponseData> object.<br>Observation: The output is aligned with the input. Similarity score: 0.71. |
| | Memory | I should save this conversation for future reference.<br>Action: storage_memory_tool<br>Action Input: < Conversation><br>Observation: The memory has been updated. |
| | Communicator | I need to ask the user in the human_tool to identify the format of the output file they prefer for the answer. Once I have that information, I can convert the provided answer into the requested format.<br>Action: human_tool<br>Action Input: What format do you prefer for the output file?<br>Observation: pdf<br>Thought: The user prefers the output file in pdf format.<br>Action: text_to_pdf_tool<br>Action Input: < Conversation><br>Observation: The PDF has been created successfully. |
| **User** | | Plot the step response for a system with transfer function num = [1, 3], den = [1, 4.16, 3.16]. |
| **LLM-Agent-Controller** | Supervisor | <Route to Memory> |
| | Memory | I need to recall if there is any previous conversation related to step response plots for transfer functions.<br>Action: recall_memory_tool<br>Action Input: Plot the step response plot for a system with transfer function num = [1, 3], den = [1, 4.16, 3.16].<br>Observation: The memory has been recalled successfully.<br>Plot the step response plot for a system with transfer function num = [1, 3], den = [1, 4.16, 3.16]. The ordered tools are tf(num, den) and step_response. TF system created is sys [46]. Step response plot successfully generated. I should start by creating a Transfer Function (TF) system from the given numerator and denominator coefficients.<br>Action: tf<br>Action Input: num = [1, 3], den = [1, 4.16, 3.16] |

| | | Observation: sys [46]<br>Inputs (1): ['u [0]']<br>Outputs (1): ['y [0]']<br>    s + 3<br>---------------------<br>s^2 + 4.16 s + 3.16<br>Thought: Now that I have the Transfer Function system, I can proceed to plot the step response.<br>Action: step_response<br>Action Input: sys = sys [46]<br>Observation:<br>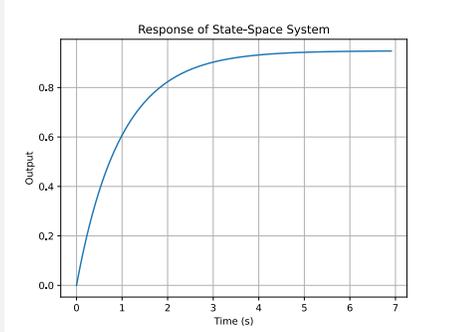<br>Thought: I now know the final answer. The step response plot for the system with transfer function num = [1, 3], den = [1, 4.16, 3.16] has been successfully generated. |
|---|---|---|
| | Communicator | I need to first understand the format the user wants the output in. Let's ask the user using the `human_tool`. After that, I will convert the answer into the appropriate format based on the user's preference.<br>Action: human_tool<br>Action Input: Please specify the format you would like the output in<br>Observation: pdf<br>Thought: The user wants the output in PDF format.<br>Action: text_to_pdf_tool<br>Action Input: < Conversation><br>Observation: The PDF has been created successfully. |